\documentclass[wcp]{jmlr}
\jmlrvolume{80}
\jmlryear{2017}
\jmlrworkshop{ACML 2017}

\title[Mutually-Dependent Hadamard Kernel]{A Mutually-Dependent Hadamard Kernel for \\Modelling Latent Variable Couplings}

\author{\Name{Sami Remes} \Email{sami.remes@aalto.fi}\\
\Name{Markus Heinonen} \Email{markus.o.heinonen@aalto.fi}\\
\Name{Samuel Kaski} \Email{samuel.kaski@aalto.fi}\\
\addr Helsinki Institute for Information Technology HIIT \\ Department of Computer Science, Aalto University}

\editors{Yung-Kyun Noh and Min-Ling Zhang}

\usepackage[utf8]{inputenc}

\usepackage{amsmath}
\usepackage{amssymb}

\usepackage{booktabs}
\usepackage{makecell}

\usepackage{capt-of}

\renewcommand{\vec}[1]{\boldsymbol{\mathbf{#1}}}
\newcommand{\mat}[1]{\vec{#1}}
\newcommand{\diag}{\operatorname{diag}}

\newcommand{\w}{\vec{w}} 
\newcommand{\z}{\vec{z}} 
\newcommand{\Z}{\mat{Z}} 
\newcommand{\y}{\vec{y}} 

\newcommand{\x}{\vec{x}} 
\newcommand{\Y}{\mat{Y}} 
\newcommand{\N}{\mathcal{N}}
\newcommand{\TN}{\mathcal{TN}}

\usepackage{tikz}
\usetikzlibrary{calc}
\usetikzlibrary{fit}

\renewcommand{\l}{{\ell}}

\renewcommand{\u}{\mathbf{u}}
\newcommand{\e}{\mathbf{e}}
\newcommand{\eps}{\boldsymbol\varepsilon}

\newcommand{\f}{\mathbf{f}}

\newcommand{\A}{\mathbf{A}}
\newcommand{\K}{\mathbf{K}}
\renewcommand{\L}{\mathcal{L}}

\newcommand{\D}{\mathcal{D}}
\newcommand{\B}{\mathbf{B}}
\newcommand{\R}{\mathbb{R}}
\newcommand{\0}{\mathbf{0}}
\newcommand{\GP}{\mathcal{GP}}
\newcommand{\GWP}{\mathcal{GWP}}

\newcommand{\bo}{\boldsymbol{\omega}}
\newcommand{\bO}{\boldsymbol{\Omega}}

\DeclareMathOperator{\cov}{\textbf{cov}}

\DeclareMathOperator*{\argmax}{arg\,max}

\DeclareMathOperator{\Tr}{Tr}
\DeclareMathOperator{\Bernoulli}{Bernoulli}
\DeclareMathOperator{\GammaD}{Gamma}

\newcommand{\pdiff}[2]{\frac{\partial #1}{\partial #2}}

\newcommand{\expt}[1]{\langle #1 \rangle}

\newcommand{\half}{\frac{1}{2}}

\begin{document}

\maketitle

\begin{abstract} 
We introduce a novel kernel that models input-dependent couplings across multiple latent processes.
The pairwise joint kernel measures covariance along inputs and across different latent signals in a mutually-dependent fashion. 
A latent correlation Gaussian process (LCGP) model combines these non-stationary latent components into multiple outputs by an input-dependent mixing matrix. Probit classification and support for multiple observation sets are derived by Variational Bayesian inference. Results on several datasets indicate that the LCGP model can recover the correlations between latent signals while simultaneously achieving state-of-the-art performance. We highlight the latent covariances with an EEG classification dataset where latent brain processes and their couplings simultaneously emerge from the model.
\end{abstract} 


\section{Introduction}

Gaussian processes (GP) are Bayesian non-parametric models that explicitly characterize the uncertainty in the learned model by describing distributions over functions \citep{rasmussen2006}. These models assume a prior over functions, and subsequently the function posterior given the data can be derived. The prior covariance plays the key roles of both regularising the model by determining its smoothness properties, and characterising how the underlying function varies in the input space.

Recently, there has been interest in deriving non-stationary covariance kernels, where the general signal variances or the intrinsic kernel parameters -- such as the lengthscales in the squared exponential or Mat\'ern kernels -- are input-dependent \citep{gibbs1997,paciorek2004,adams2008,tolvanen2014,heinonen2016}. For instance, in geostatistical applications, a non-stationary kernel can both model a difference in the covariance along or across geological formations \citep{goovaerts1997}. Input-dependent, heteroscedastic noise models have also been studied in single-task \citep{goldberg1998,le2005,kersting2007,quadrianto2009,lazaro2011,wang2012} and in multi-task settings \citep{rakitsch2013}. 

In multi-task learning Gaussian processes are utilized by modeling the output covariances between possibly several latent functions \citep{yu2005,bonilla2007,alvarez2010,alvarez2012}. In latent function models\footnote{Coined as \emph{linear models of coregionalisation} (LCM) in geostatistics literature \citep{goovaerts1997}.} the outputs are linear combinations of multiple underlying latent functions \citep{seeger2005,schmidt2009}. In Gaussian Process Regression Networks (GPRN) the mixing coefficients of multiple independent latent signals are input-dependent Gaussian processes as well, leading to a general multi-task framework that adaptively combines latent signals into outputs along the input space \citep{wilson2012}.

The main contribution of this paper is to introduce a mutually-dependent Hadamard product kernel that combines a covariance structure between the latent signals that depends on the inputs, with an input kernel that depends on the latent signal indices. The signal and input kernels are interdependent, conditional on each other. This is in contrast to earlier Kronecker-based joint kernels where inputs and latent signals would be assumed independent. The kernel generalizes Wishart processes \citep{wilson2011} into cross-covariances for input-dependent correlation structure, and a non-stationary Gaussian kernel \citep{gibbs1997} for measuring input-space correlations at specific latent signals. We deploy this kernel to extend the GPRN framework by a non-stationary cross-covariance function for the latent signals.

Furthermore, the proposed latent correlation Gaussian process (LCGP) incorporates multiple latent signals that are linearly combined into multiple outputs in an input-dependent fashion. The latent signals have a structured Wishart-Gibbs model that leads to non-stationary signal variances. We account for both regression, and Probit-based classification. Finally, the model is extended for multiple observation sets, where each observation is modeled by a separate latent model with shared latent correlations. In such a model, the latent correlations effectively regularize the latent models of each observation. Variational Bayesian inference with whitened gradients is derived for a scalable implementation. 

We highlight the model with several datasets where interesting latent signal covariance models emerge, while retaining or improving the state-of-the-art regression and classification performance. Multi-observation classification is demonstrated on EEG data from a large set of scalp measurements from several subjects, where the model is able to learn the covariance model between the underlying brain processes. In simulation studies, we show that our model is capable of accurately learning the latent variable correlations.

\section{Latent Correlation Gaussian Process}

We consider $M$-dimensional observations $\y(x) \in \R^M$ over $N$ data points $(x_1, \ldots, x_N)$. We denote vectors with boldface symbols, matrices with capital symbols and block matrices with boldface capital symbols. In this section we first construct the multi-output regression model for $\y(x)$, and then develop a novel kernel for latent variables in such a model as our main contribution. Section~\ref{sec:classif} further extends the framework into a classification setup.

\subsection{Multi-output regression}

Following \citet{wilson2012}, we model the $M$-dimensional outputs $\y(x) \in \R^M$ as an input-dependent mixture of $Q$ latent signals $\u(x) \in \R^Q$ via a mixing matrix $B(x) \in \R^{M \times Q}$,
\begin{align}
\y(x) &= \f(x) + \e = B(x)\big(\u(x) + \eps\big) + \e, \label{eq:likelihood}
\end{align}
where $\e = \e(x)$ is zero-mean $M$-dimensional Gaussian observation noise and $\eps = \eps(x)$ is zero-mean $Q$-dimensional latent noise
\begin{align*}
\e &\sim \N(\0, \omega_f^{-1} \vec I), \quad \omega_f \sim \GammaD( \alpha_f, \beta_f), \\
\eps &\sim \N(\0, \omega_u^{-1} \vec I),  \quad \omega_u \sim \GammaD(\alpha_u, \beta_u) \, .
\end{align*}

We model both the latent variables $\u$ as well as the elements of the mixing matrix $B$ as Gaussian processes. A GP prior $\phi(x) \sim \GP(\mu(x),k(x,x'))$ defines a distribution over functions $\phi(x)$ with expectation $\mathbb{E}[\phi(x)] = \mu(x)$ and covariance of the values between points $x$ and $x'$ is $\cov [\phi(x),\phi(x')] = k(x,x')$. A set of function values $\vec\phi = (\phi(x_1), \ldots, \phi(x_N))^T$ follows a Gaussian $\vec\phi \sim \N(\boldsymbol\mu, K)$ with $K_{ij} = k(x_i,x_j)$ and $\mu_i = \mu(x_i)$.

The mixing matrix $B(x)$ is an $M \times Q$ matrix of independent Gaussian processes over outputs $m$ and latent signals $q$,
\begin{align}
B_{mq}(x) \sim \GP(0, k_b(x,x')).
\end{align}
The kernel $k_b(x,x')$ between two input points $x$ and $x'$ determines how mixing of latent signals into outputs evolves along the input space. For instance, with temporal data the mixing matrix allows time-dependent linear combinations of the outputs. The full model is depicted in Figure~\ref{fig:gm}. Next, we proceed to derive a kernel for the latent variables $\u$.

\subsection{Wishart-Gibbs Hadamard Product Kernel}

\begin{figure}[t]
    \centering
    \includegraphics[width=0.9\columnwidth]{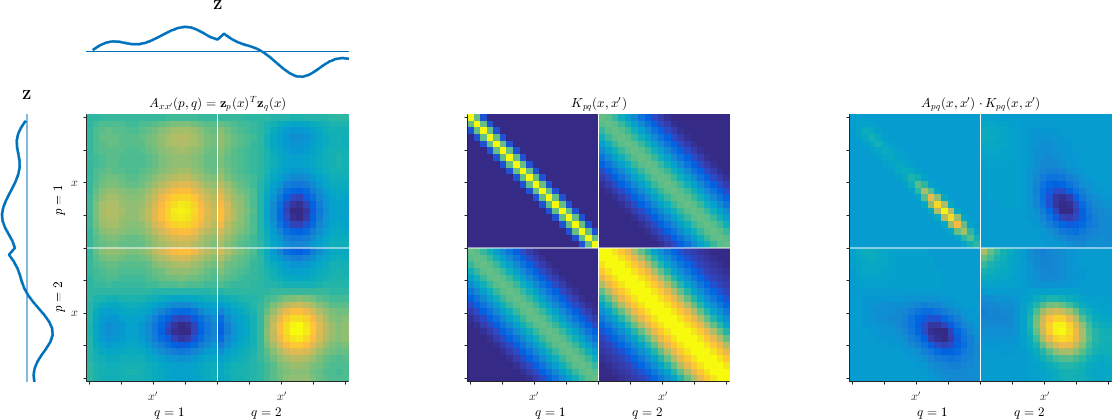}\vspace{.1cm}
    \hspace*{.75cm}
    \includegraphics[width=0.85\columnwidth]{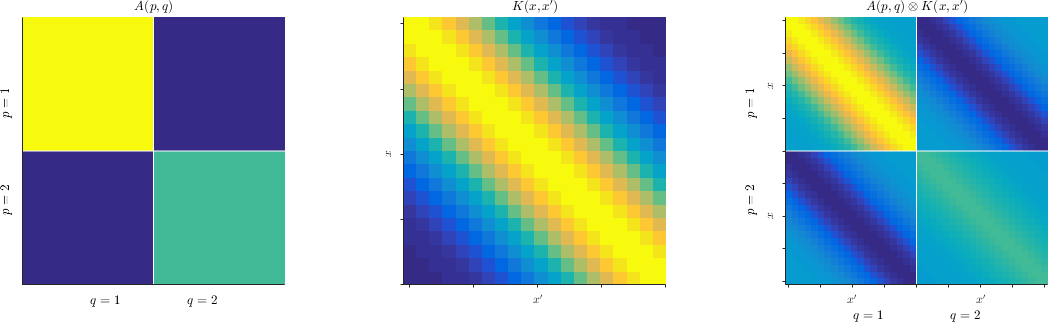}
    \caption{Illustration of the proposed kernel. Top: The input-dependent signal kernel $A_{xx'}(p,q)$ and signal-dependent input kernel $K_{pq}(x,x')$ are mixed into a rich and flexible pairwise combined kernel. Bottom: A standard Kronecker kernel between signals and inputs can not reproduce dependent correlations.}
    \label{fig:lck}
\end{figure}

The latent signals $u_p(x)$ are functions of the signal index $p$ and input $x$. We propose to encode the latent signals $\u(x)$ as \emph{mutually dependent} Gaussian processes over pairs of inputs $(x, x')$ and signals $(p, q)$,
\begin{align}
u_p(x) &\sim \GP\left(0, A_{xx'}(p,q) K_{pq}(x,x') \right),
\end{align}
such that the joint covariance $\cov[ u_p(x), u_q(x')] = A_{xx'}(p,q) K_{pq}(x,x')$ is a product of signal and input similarities. Both similarities depend on each other to produce a non-stationary joint covariance.

The pairwise, mutually dependent Hadamard kernel $k(x,x',p,q) = A_{xx'}(p,q) K_{pq}(x,x')$ encodes a rich similarity between input $x$ of latent signal $p$ and $x'$ of latent signal $q$ as the product of the two conditional kernels. The kernel $A_{xx'}(p,q)$ encodes signal similarity between inputs $x$ and $x'$, while the kernel $K_{pq}(x,x')$ denotes input similarity at latent signals $p$ and $q$. Since the two kernels depend on each other, a simple model such as Kronecker kernel product \citep{stegle2011} is not suitable. Both kernels can be interpreted as cross-covariances. The Gibbs kernel $K_{pq}$ restricts the flexibility of the Wishart kernel $A_{xx'}$ (See Figure \ref{fig:lck}).

For instance, in EEG data the kernels could signify correlations $A_{tt'}(p,q)$ between latent brain processes $p$ and $q$ at two time points $t$ and $t'$, while $K_{pq}(t,t')$ is a smooth temporal kernel that connects events that occur at similar time points. In geospatial applications, the correlations $A_{\x \x'}(p,q)$ can encode similarity between two latent ore functions $p$ and $q$ at two locations $\x,\x' \in \R^2$, for instance between cadmium and zinc concentrations \citep{goovaerts1997}. The location kernel $K_{pq}(\x, \x')$ could encode a smooth spatial proximity. A conventional Kronecker kernel $k(\x,\x',p,q) = A(p,q) K(\x,\x')$ would assume -- in contrast -- that (i) the same spatial proximity $K(\x,\x')$ applies to all ore functions $p$, and (ii) two ore concentrations would correlate similarly independent of the location $\x$. 

We start forming the joint kernel by considering a non-stationary Gaussian kernel for the inputs $x$  \citep{gibbs1997},
\begin{align}
K_{pq}(x,x') = \sqrt{\frac{2\ell_p\ell_q}{\ell_p^2+\ell_q^2}}\exp\left(-\frac{(x-x')^2}{\ell_p^2+\ell_q^2}\right),
\end{align}
which encodes specific lengthscales $\l_1, \ldots, \l_Q$ for each latent signal. 
The kernel within a single latent signal $K_{pp}$ reduces into a standard Gaussian kernel, while the cross-covariance similarity $K_{pq}$ measures similarity of two inputs with different associated lengthscales.

We base our construction of the mutually dependent covariance structure $A_{xx'}(p,q)$ on Wishart processes. A Generalized Wishart Process (GWP) prior on a covariance matrix, that depends on a single variable $x$, is \citep{wilson2011}
\begin{align*}
A(x) &= \sum_{r=1}^\nu L \z_r(x) \z_r(x)^T L^T \sim \GWP(V,\nu,K_z),
\end{align*}
where $V = LL^T$ and all $z_{pr}(x) \sim \GP(0, K_z(x,x') )$
are independent Gaussian processes for $p = 1, \ldots, Q$ and $r = 1, \ldots, \nu$. 
The kernel $K_z$ determines the change of $A(x)$ in the input space. From this formulation we define our joint kernel, such that we preserve the GWP marginal for $A(x)$ by extending the GWP into \emph{cross-covariances} of two variables, as
\begin{align}
A_{xx'}(p,q) = \z_p(x)^T \z_q(x'),
\end{align}
where we have for each element of $\z_p(x) \in \R^\nu$ a GP prior (See Figure \ref{fig:lck}). With this choice the prior expectation of the covariance is the identity matrix.

The resulting covariance of $u_p(x)$ is then a product of covariance between inputs at signals $p$ and $q$, and a covariance between signals at inputs $x$ and $x'$. 
This covariance can be seen marginally from two perspectives, 
\begin{align*}
\u_p &\sim \N_N(\0, A(p,p) \circ K_{pp}) \\
\u(x) &\sim \N_Q(\0, A(x) \circ P),
\end{align*}
where $\u_p \in \R^N$ is a single latent signal that follows a Normal distribution weighted by variances $A(p,p)$, and $\u(x) \in \R^Q$ contains all $Q$ latent signals at input $x$ and follows a Normal distribution with generalized Wishart process prior, scaled by the matrix $P_{pq} = \sqrt{\frac{2\ell_p\ell_q}{\ell_p^2+\ell_q^2}}$. The element-wise, or Hadamard, product of $x$ and $y$ is denoted by $x \circ y$.

The joint covariance over the concatenated column vector of all latent signals $\u \in \mathbb{R}^{QN}$ is a block matrix
\begin{align*}
\cov(\u, \u) &= \left( Z_i Z_j^T \circ K_{ij} \right)_{i,j=1}^N + \bO_u = \Z \Z^T \circ \K_Q + \bO_u 
\end{align*} 
where $Z_i$ is a $Q \times \nu$ matrix, and $\Z = (Z_1, \ldots, Z_N)^T$. 
The $Q \times Q$ kernel $K_{ij} = (K_{pq}(x_i,x_j))_{p,q=1}^Q$ gives the signal similarities at inputs $x_i$ and $x_j$, the block matrix $\K_Q = (K_{ij})_{i,j=1}^N$ collects them into $(N \times N)$ blocks of kernel values, and finally the noise matrix is $\bO_u = \omega_u^{-1} \vec I_{QN}$, introducing the latent noise directly into the covariance of the $\u$'s. The resulting joint input-output covariance $\Z\Z^T \circ \K_Q$ consists of $N \times N$ block matrices of size $Q \times Q$. See Figure~\ref{fig:lck} for a visualisation. 
The kernel matrix $\A_Q$ is positive semi-definite (PSD) as an outer product, and the Gaussian kernel is PSD as well \citep{gibbs1997}. The Hadamard product $\A_Q \circ \K_Q$ retains this property. We refer to this kernel as the Wishart-Gibbs cross-covariance (WGCC).

The proposed latent correlation Gaussian process (LCGP) model is a flexible Bayesian regression model that simultaneously infers the latent signals and their mixing to match the output processes, while learning the underlying correlation structure of the latent space using the WGCC kernel. The latent correlations are parameterised by two terms that characterise the input and signal similarities with Gaussian and Wishart functions, respectively. A key feature of the model is the ability adaptively couple and decouple latent processes along the input space.

\section{Classification with multiple observations}\label{sec:classif}

We further suppose that we have $S$ observations or samples $\y^{(s)}(x)$ associated with a class label, or response, $r^{(s)}$, and assume that all these observations share their latent space. We then learn separate latent functions $\u^{(s)}$ for each sample, while keeping the mixing model $B(x)$, latent correlations $\A(x)$ and $\K_Q$, and the noise precisions $\omega_f$ and $\omega_u$ shared. The noiseless sample is then reconstructed as $$\f^{(s)}(x) = B(x) \u^{(s)}(x),$$
which results in the same likelihood as in eq.~\eqref{eq:likelihood}.

We build a classifier in the latent signal space as
a Probit classification model over all latent signals $\w^T \u^{(s)}$ with Gaussian-Gamma priors, where $\w \in \R^{NQ}$ is a concatenated column vector of linear weights $\w_p \in \R^N$ for the $Q$ latent signals. This allows us to reduce the data dimensionality for classification, as $M$ can potentially be very large. The classifier is then 
\begin{align}
    r^{(s)} \mid \w, \u^{(s)} &\sim \Bernoulli(\Phi(\w^T \u^{(s)} + b)), \label{eq:classlik} \\
    \w_p\mid\lambda_w &\sim \N(0, \lambda_w^{-1}), \, \lambda_w \sim \GammaD( \alpha_w, \beta_w) \nonumber\\
    b\mid\lambda_b &\sim \N(0, \lambda_b^{-1}), \, \lambda_b \sim \GammaD(\alpha_b, \beta_b), \nonumber
\end{align}
where we index the observations with $s$, and $\w$ and $b$ are the classifier weights and bias, respectively. The Gaussian CDF is denoted by $\Phi(\cdot)$. We additionally assume, for notational clarity, that all data are observed at the same input points $x_1, \ldots, x_N$.  

Essentially, our model now has two likelihoods for the two types of data, one defined for the output data in eq.~\eqref{eq:likelihood} and one for the class labels related to the outputs in eq.~\eqref{eq:classlik}.

\section{Inference}

\subsection{Variational Bayes}
For inference in our Bayesian model we adopt the Variational Bayesian (VB) approach \citep{attias1999inferring}, which is
based on maximising a lower bound on the log marginal likelihood of the data with respect to a distribution $q(\Theta)$, where $\Theta$ represents all model parameters. The lower bound is of an easier form than the true posterior distribution $p(\Theta|\D)$, where $\D = (Y^{(s)},r^{(s)})_{s=1}^S$ and $Y^{(s)} \in \R^{M \times N}$. The lower bound is obtained
by Jensen's inequality
\begin{align*}
    \log p(\D) &= \log \int q(\Theta) \frac{p(\D, \Theta)}{q(\Theta)} d\Theta \geq \int q(\Theta) \log \frac{p(\D, \Theta)}{q(\Theta)} d\Theta \equiv \L(q) \, .
\end{align*}
Typically, a factorised approximation $q(\Theta) = \prod_i q(\theta_i)$
is used, where $\theta_i$ are some disjoint subsets of the variables $\Theta$. It can be
shown that the optimal solution that maximizes $\L(q)$ is
\[ q(\theta_i) \propto \exp(\langle \log p(\Theta, \D) \rangle_{\theta_{-i}}), \]
in which the expectation is taken with respect to all variables except $\theta_i$.
The VB algorithm consists of iterating through updating each factor $q(\theta_i)$.

\subsection{VB for the LCGP classification model}

We employ the following factorization 
\begin{align}
    q(\Theta) &= \prod_s q(\u^{(s)}) q(h^{(s)}) \prod_m q(\B_m) q(\omega_f) q(\w,b) q(\lambda_w) q(\lambda_b) q(\Z), 
\end{align}
where $q(\B_m)$ factorizes the mixing matrix $\B$ row-wise.
Most factors have standard distributions, the update formulas are shown in Table~\ref{tab:vb-updates}. The VB inference procedure is summarised in Algorithm~\ref{algo:vb}. LCGP can be run with or without the classification part of the model; without classification the parameters involved are ignored (see Figure~\ref{fig:gm}).

\begin{table}[t]
    \floatconts{tab:vb-updates}{\caption{VB updates. Here $\tilde\u$ collects $\u(x_i)$ into a block diagonal matrix, and $q(\lambda_b)$ is analogous to $q(\lambda_w)$. Auxiliary variables $h$ have a Gaussian posterior truncated either to negative or positive values depending on $r^{(s)}$.}}%
    {\begin{tabular}{ll}
    \toprule
    Distribution $q(\cdot)$ & Parameters \\
    \midrule
    $ q(\u^{(s)}) = \N(\u^{(s)} \mid \vec\mu_u^{(s)}, \vec\Sigma_u) $ & \makecell[cl]{$\vec\Sigma_u^{-1} = (\Z\Z^T \circ \K_Q + \bO)^{-1} + \expt{\omega_f}\expt{\B^T\B} + \expt{\w\w^T}$ \\ $\vec\mu_u^{(s)} = \vec\Sigma_u \left( \left[\expt{h^{(s)}}-\expt{b}\right]\expt{\w} + \expt{\omega_f}\expt{\B^T}\y^{(s)} \right)$} \\
    $q(\B_m) = \N(\B_m \mid \vec\mu_m, \vec\Sigma_b)$ & \makecell[cl]{$\vec\Sigma_b^{-1} = \K_b^{-1} + \expt{\omega_f}\sum_s \expt{\tilde\u^{(s)T}\tilde\u^{(s)}}$ \\ $\vec\mu_m = \vec\Sigma_b \expt{\omega_f} \sum_s \expt{\tilde\u^{(s)T}\y_m^{(s)}}$} \\
    $q(\w,b) = \N\left( \begin{pmatrix} \w \\ b \end{pmatrix} \mid \vec\mu_{w,b}, \vec\Sigma_{w,b} \right)$ & \makecell[cl]{$\vec\Sigma_{w,b}^{-1} = \begin{pmatrix} \expt{\u\u^T} + \diag\expt{\lambda_w} & \expt{\u}\vec1 \\ \vec1^T\expt{\u}^T & S+\expt{\lambda_b} \end{pmatrix}$ \\ $\vec\mu_{w,b} = \vec\Sigma_{w,b} \begin{pmatrix} \expt{\u}\expt{\vec h} \\ \vec1^T\expt{\vec h} \end{pmatrix}$} \\
    $q(\lambda_w) = \GammaD(a_w, b_w)$ & $a_w = \alpha_w + \half NQ, \quad b_w = \beta_w + \half \expt{||\w||^2}$ \\
    $q(\omega_f) = \GammaD(a_{\omega_f}, b_{\omega_f})$ & \makecell[cl]{$a_{\omega_f} = \alpha_f + \half NMS, \quad \beta_f + \half \expt{||\y-\B\u||^2_2}$} \\
    $q(h^{(s)}) = \mathcal{TN}_{r^{(s)}}(h^{(s)} \mid g^{(s)}, 1)$ & \makecell[cl]{$g^{(s)} = \expt{\w^T}\expt{\u^{(s)}}+\expt{b}$} \\
    \bottomrule
    \end{tabular}}
\end{table}

Auxiliary variables $h$ are introduced to make the variational inference tractable for Probit classification \citep{albert1993bayesian},
\begin{align}
    h \mid \w, \u \sim \N(\w^T\u + b, 1). 
\end{align}
Class labels depend on the sign of $h$, i.e. $r = +1$ if $h > 0$. Integrating out $h$ recovers the Probit likelihood 
\begin{align}
    p(r|\w,\u) = \Bernoulli( r | \Phi(\w^T\u+b)).
\end{align}
The posterior $q(h)$ is a truncated Gaussian \citep{albert1993bayesian}, which has analytical formulas for first and second moments.

Finally, $q(\Z)$ is updated by optimising the lower bound $\L(\Z)$ with respect to $\Z$. We optimise $\L(\Z)$ using L-BFGS in whitened domain employing a change of variables $\hat\Z = \vec L^{-1}\Z$ with the Cholesky of the kernel $\K_z = \vec L \vec L^T$ to make the optimization more efficient \citep{kuss2005,heinonen2016}, see the Supplementary for details.

Predictions to new inputs $x^*$ can be made by applying standard GP formulas to obtain $\B(x^*)$, $\u(x^*)$ and $\Z(x^*)$ based on the optimized variational posterior $q(\Theta)$. For new observation with unkown class label $r^{(s^*)}$, we can apply the update for $q(\u^{(s^*)})$ without classification related terms.

\begin{figure}
    \begin{minipage}{0.50\textwidth}%
    \centering
     \resizebox{\textwidth}{!}{%
     \begin{tikzpicture}
     \tikzstyle{roundnode}=[circle, minimum size = 7.5mm, thick, draw =black!80, node distance=14mm,align=center,text width=7.5mm,inner sep=0pt]
      \tikzstyle{roundednode}=[rounded corners=3pt, minimum size=8mm, thick, draw=black!80, node distance=13mm]
      \tikzstyle{opennode}=[minimum size = 8mm, thick, node distance=13mm]
     \tikzstyle{connect}=[-latex, thick]
     \node[roundnode,fill=black!5] (Y) [] {$Y^{(s)}$};
     \node[roundnode] (of) [above of=Y] {$\bo_f$};
     \node[roundnode] (B) [left of=of] {$\B$};
     \node[opennode] (Kb) [above of=B] {$K_b$};
     \node[roundnode] (U) [right of=of] {$\u^{(s)}$};
     \node[roundnode] (A) [above of=U, label={[shift={(1.0,0.1)}]$\A=\Z\Z^T$}] {$\A$};
     \node[opennode] (Kz) [above of=A] {$K_z$};
     \node[roundnode] (ou) [left of=A] {$\bo_u$};
     \node[opennode] (Ku) [right of=A] {$K_{pq}$};
     \node[roundnode] (lw) [right of=Ku] {$\lambda_w$};
     \node[roundnode] (lb) [right of=lw] {$\lambda_b$};
     \node[roundnode] (w) [below of=lw] {$\w$};
     \node[roundnode] (b) [below of=lb] {$b$};
     \node[roundnode,fill=black!5] (r) [below of=w] {$r^{(s)}$};
     \path 
        (B) edge [connect] (Y)
        (Kb) edge [connect] (B)
        (of) edge [connect] (Y)
        (U) edge [connect] (Y)
        (U) edge [connect] (r)
        (ou) edge [connect] (U)
        (A) edge [connect] (U)
        (Kz) edge [connect] (A)
        (Ku) edge [connect] (U)
        (w) edge [connect] (r)
        (b) edge [connect] (r)
        (lw) edge [connect] (w)
        (lb) edge [connect] (b);
        
   \node[yshift=16pt, inner sep=8pt, inner ysep=20pt,fit=(lw) (lb) (r)] [label={[shift={(-0.30,-0.5)}] Classification}] {};
   \node[yshift=-3pt,inner xsep=4pt, inner ysep=2pt, fit=(Kb) (Kz) (Ku) (Y)] [label={[shift={(-1.95,-0.55)}] LCGP}] {};
   \node[draw,rounded corners=.15cm,fit=(Kb) (Kz) (Y) (lb)] [] {};
   \draw [dashed,very thick,black!80] (3.25,-0.35) -- (3.25,4.6);
    \end{tikzpicture}
    }
    \captionof{figure}{Graphical model of the LCGP.}
    \label{fig:gm}
    \end{minipage}
\hfill
\begin{minipage}{.45\textwidth}%
\hspace{-12pt}
\begin{algorithm2e}[H]
    \small
    \textbf{Input}: Data $\Y$ and $\vec r$, kernel parameters, initialized $q(\cdot)$ \\
    \While{not converged}{
    $q(\u^{(s)}) \gets \N(\vec\mu_u^{(s)},\vec\Sigma_u)\,\forall s$ \\
    $\Z,\,\omega_u \gets \argmax_{\Z,\omega_u} \L(\Theta)$ \\
    \If{do classification}{
    $q(h^{(s)}) \gets \TN(g^{(s)},1) \, \forall s$ \\
    $q(\w,b) \gets \N(\vec\mu_{w,b},\vec\Sigma_{w,b})$ \\ $q(\lambda_w) \gets \GammaD(a_w,b_w)$
    }
    $q(\B_m) \gets \N(\vec\mu_m, \vec\Sigma_b) \, \forall m$ \\
    $q(\omega_f) \gets \GammaD(a_{\omega_f}, b_{\omega_f})$
    }
    \caption{VB for LCGP.}
    \label{algo:vb}
\end{algorithm2e}
\end{minipage}
\end{figure}

\section{Related Work}

In semiparametric latent factor models (SLFM) the signal $\f(x) = B \u(x)$ over $M$ outputs is a linear combination of $Q$ independent latent Gaussian process signals $\u(x)$ with a fixed mixing matrix $B \in \R^{M \times Q}$, with appropriate hyperparameter learning \citep{seeger2005}. A Gaussian process regression network (GPRN) \citep{wilson2012} extends this model by considering a mixing matrix $B(x)$ where each element $B_{pi}(x)$ is an independent Gaussian process along $x$.

In geostatistics vector-valued regression with Gaussian processes is called \emph{cokriging} \citep{alvarez2012}. In \emph{linear coregionalization models} (LCM) latent Gaussian processes are mixed from latent signals $u_p(x)$ and $u_q(x')$ that are independent. In contrast to SLFM, each signal $u_p(x)$ is an additional mixture of $R_Q$ signals with separate shared covariances $K_q(x,x')$. In the intrinsic coregionalization model (ICM) only a single ($Q=1$) latent mixture with a single shared kernel is used, while in SLFM there are multiple latent singleton ($R_Q=1$) signals. In spatially varying LCMs (SVLCM) the mixing matrices are input-dependent, similar to GPRNs \citep{gelfand2004}. \citet{vargas2002} used non-orthogonal latent signals $u_p(x)$ and $u_q(x')$ with fixed covariances.

Multi-task Gaussian processes employ structured covariances that combine a task covariance with an input covariance. Simple Kronecker products between the covariances assume that task and input covariances are independent functions  \citep{bonilla2007,stegle2011,rakitsch2013}. This is computationally efficient \citep{flaxman2015}, but it does not take into account interactions between the tasks and inputs.

In Generalised Wishart Processes \citep{wilson2012} an input-dependent covariance matrix $\Sigma(x) = \sum_{n=1}^\nu  \z_n(x) \z_n(x)^T$ is a sum of $\nu$ outer products. 
The random variables $z_{ni}(x) \sim \GP(0, K(x,x'))$ are all independent Gaussian processes. Copula processes also describe dependencies of random variables by Gaussian processes \citep{wilson2010}. In Bayesian nonparametric covariance regression, covariances of multiple predictors share a common dictionary of Gaussian processes \citep{fox2015}.

Finally, Gaussian process dynamical or state-space systems are a general class of discrete-time state-space models that combine the latent state into time-dependent outputs as Markov processes \citep{wang2005,damianou2011,deisenroth2012,frigola2014}. In Gaussian process factor analysis the outputs are described as factors that have GP priors, however not modeling the factor dependencies \citep{lawrence2004,luttinen2009}.

\section{Experiments}

In the first experiment we show that our model\footnote{Our Matlab implementation can be found on \url{https://github.com/sremes/wishart-gibbs-kernel}} can recover the true latent correlations in a simple simulated-data case, and compare our method with GPRNs, which is a state-of-the-art multi-output Gaussian process regression model \citep{wilson2012}. We employ the mean-field variational inference implementation of GPRN by \citet{nguyen2013efficient}. Second, we apply our method to the Jura geospatial dataset to elucidate latent ore concentration process couplings. Finally, we demonstrate our full modelling framework on an EEG single-trial classification task, outperforming state-of-the-art regularised LDA in classification and additionally recovering an interesting latent representation that we further evaluate in a simulation study.
Results from the experiments are summarised in Table~\ref{tab:results}.

\begin{table}[t]
    \caption{Results on all datasets. Boldface numbers indicate better method performance. MAE and MSE refer to the mean absolute and squared errors, respectively. AUC refers to the area under the ROC curve statistic. For the EEG simulation study, the Fisher's method is used to combine p-values from the simulations.} \label{tab:results}
    \begin{center}
    \begin{tabular}{lrr}
    \hline
    \texttt{JURA} & Average MAE & Average MSE \\
    \hline
    \, LCGP & \textbf{0.686} $\pm$ 0.057 & 0.804 $\pm$ 0.16\\
    \, GPRN & 0.693 $\pm$ 0.033 & \textbf{0.801} $\pm$ 0.09 \\
    \hline
    \texttt{EEG: classif.} & Average AUC$^1$ & \\
    \hline
    \, LCGP & \textbf{0.830} $\pm$ 0.0022 & \\
    \, RLDA & 0.826 $\pm$ 0.0020 & \\
    \hline
    \texttt{EEG: simulation} & Mean score & $p$-value \\
    \hline
    \, $Q=2$ & 0.87 & 9.62e-10 \\
    \, $Q=3$ & 0.84 & 4.21e-08 \\
    \, $Q=4$ & 0.87 & 1.11e-15 \\
    \end{tabular}
    \end{center}
    \vspace{-.5em}
    {\hspace{8em}\footnotesize $ ^1$ Paired t-test, $p < 0.05$.}
\end{table}

\subsection{Simulated Data Experiments}

\subsubsection{Wishart-Gibbs kernel in multi-output GP}

\begin{figure}
    \centering
    \includegraphics[width=.85\textwidth]{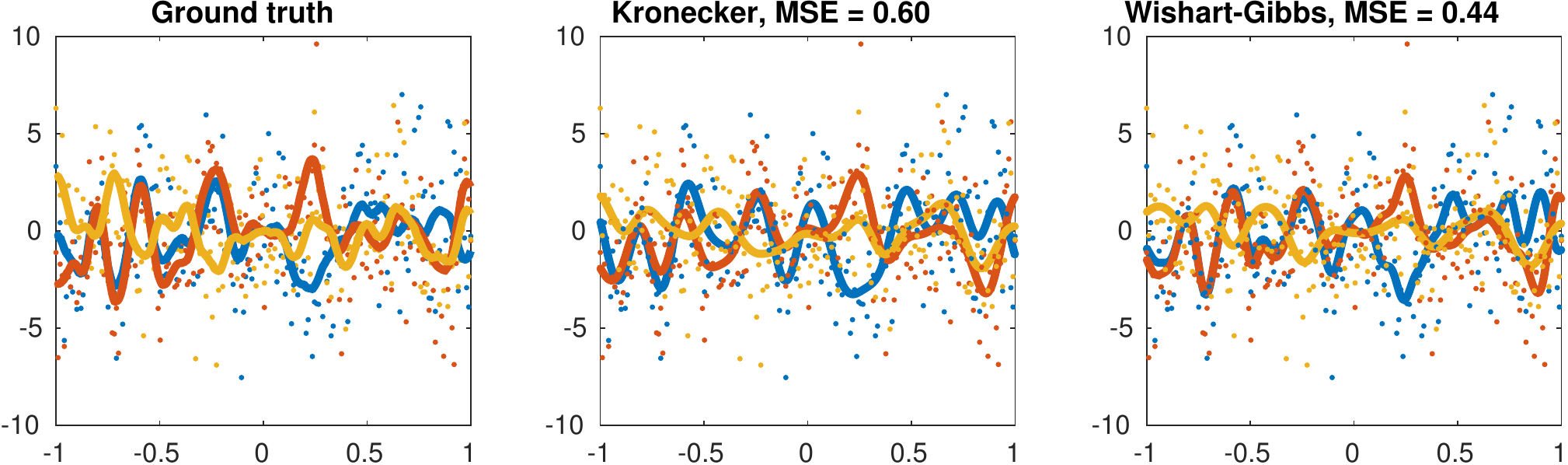}
    \caption{A noisy three-output simulated dataset with high coupling between the outputs that changes at the origin (see the blue and red lines). The proposed Wishart-Gibbs kernel obtains much better match to the ground truth.}
    \label{fig:kron-vs-hadam}
\end{figure}

We simulated a dataset that contains a clear switch in the coupling of three outputs in the middle of an interval $[-1, 1]$. A traditional Kronecker multi-output kernel of form $\K = A \otimes K$, with $A = \sum_k \z_k\z_k^T$ and $K$ a Gaussian kernel, cannot model this, but our proposed Wishart-Gibbs kernel can adapt to this switch point. The data and posterior fit with both kernels are shown in Figure~\ref{fig:kron-vs-hadam}. Our kernel obtains an MSE of \textbf{0.44}, and Kronecker kernel 0.60, with a baseline of 1.92 (predicting zero).

\subsubsection{Recovering latent covariance with LCGP}
We use simple toy data to show that we are able to recover known latent correlation structure.
We generated data with a varying number of latent components $Q = 2, \ldots, 5$ and amount of samples $S = 1, \ldots, 20$. The mixing matrix was binary such that one output maps to one latent variable. For simplicity, we only consider LCGP without the classifier.

\begin{figure}
    \centering
    \includegraphics[width=.8\linewidth]{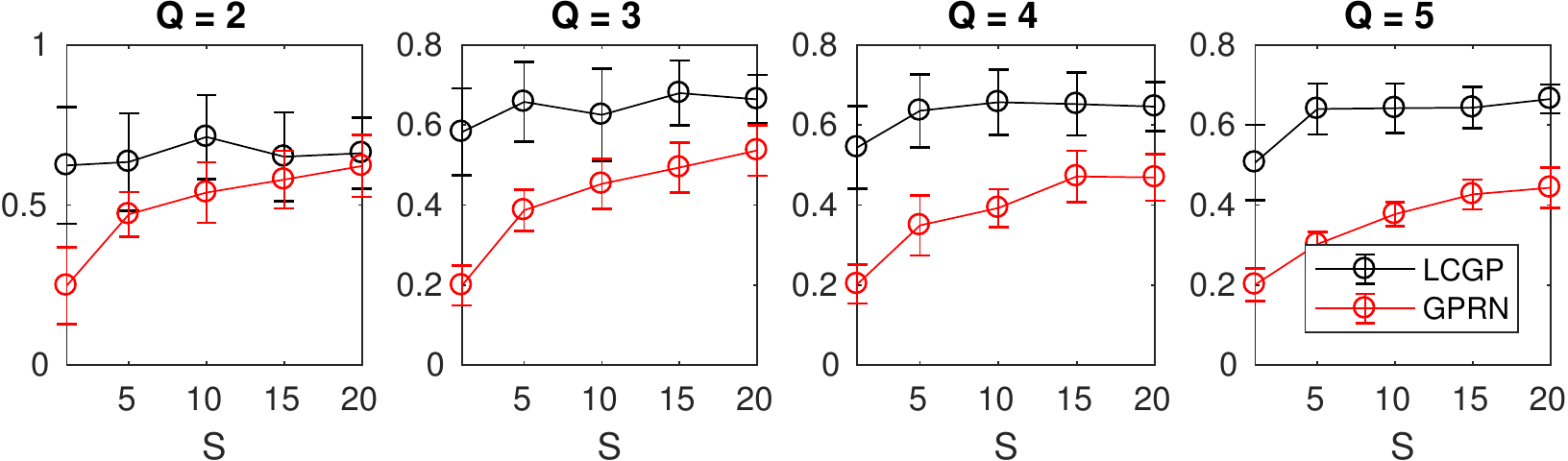}
    \captionof{figure}{The true latent covariances can be recovered more accurately with our method than with GPRN. The curves show correlation to the true correlation matrix, over the elements of the recovered correlation matrix, as a function of the (very small) sample size $S$. The $Q$ is the number of latent signals.}
    \label{fig:toy}
\end{figure}

To assess the accuracy, we measured the correlation between the elements of the true covariance matrix to the one estimated. With GPRN we computed the empirical covariance $\hat{\mat\Sigma} = \sum_s \u^{(s)}\u^{(s)} \,^T$ of the latent variables. As the order of the recovered latent variables is not identifiable, we computed the correlations over all permutations and report the best. Rotations of the latent space are not accounted for, however. The results in  Figure~\ref{fig:toy} show that our model can recover the true underlying latent covariance with high correlations.

\subsection{Jura}

\begin{figure}[t]
    \centering
    \subfigure[$ \cov(u_i(\x),u_j(\x)) $]{\includegraphics[height=2.2in]{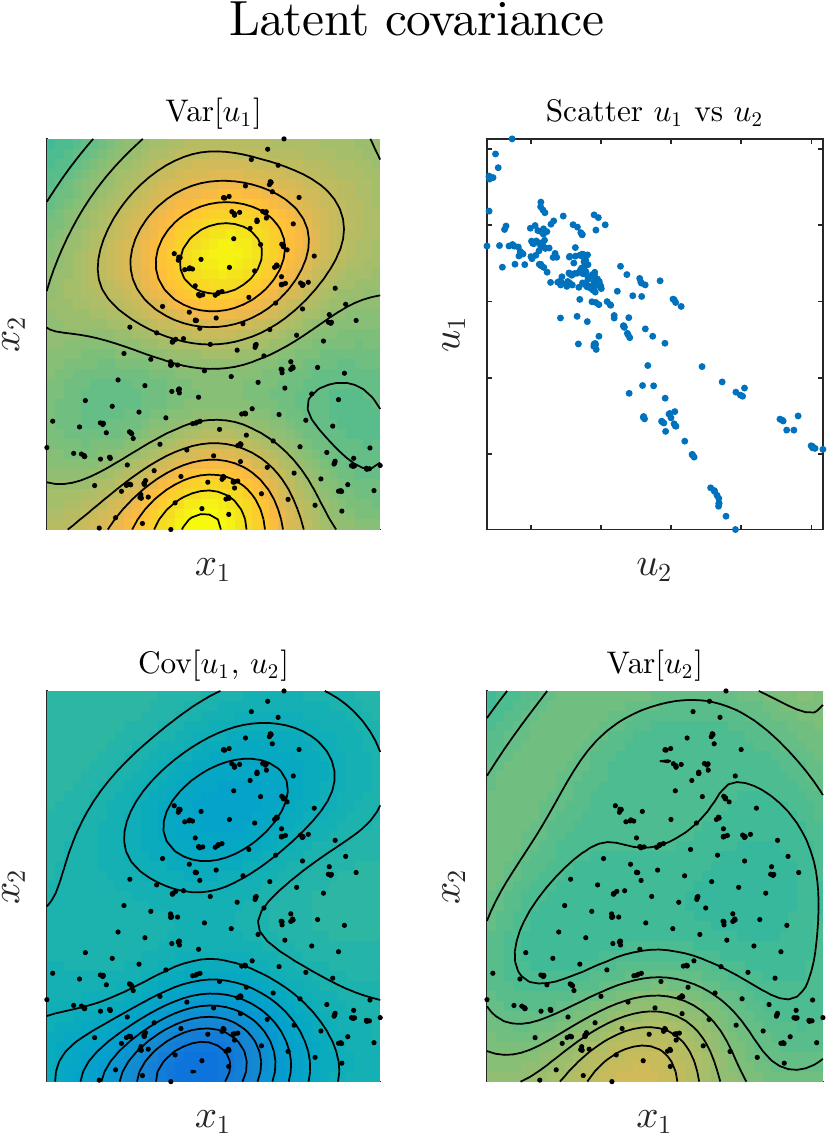}}
    \hfill
    \subfigure[$u_i(\x)$]{\includegraphics[height=2.2in]{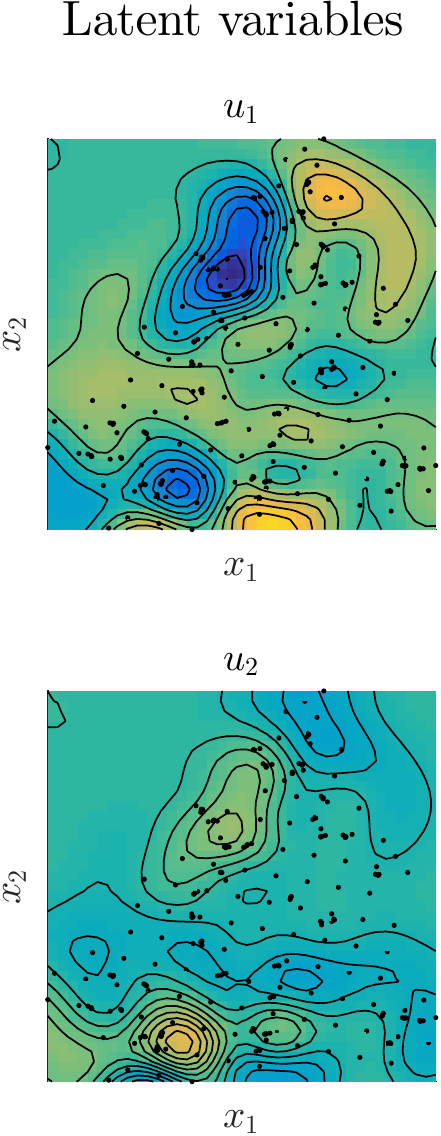}}
    \hfill
    \subfigure[$\B(\x)$]{\includegraphics[height=2.2in]{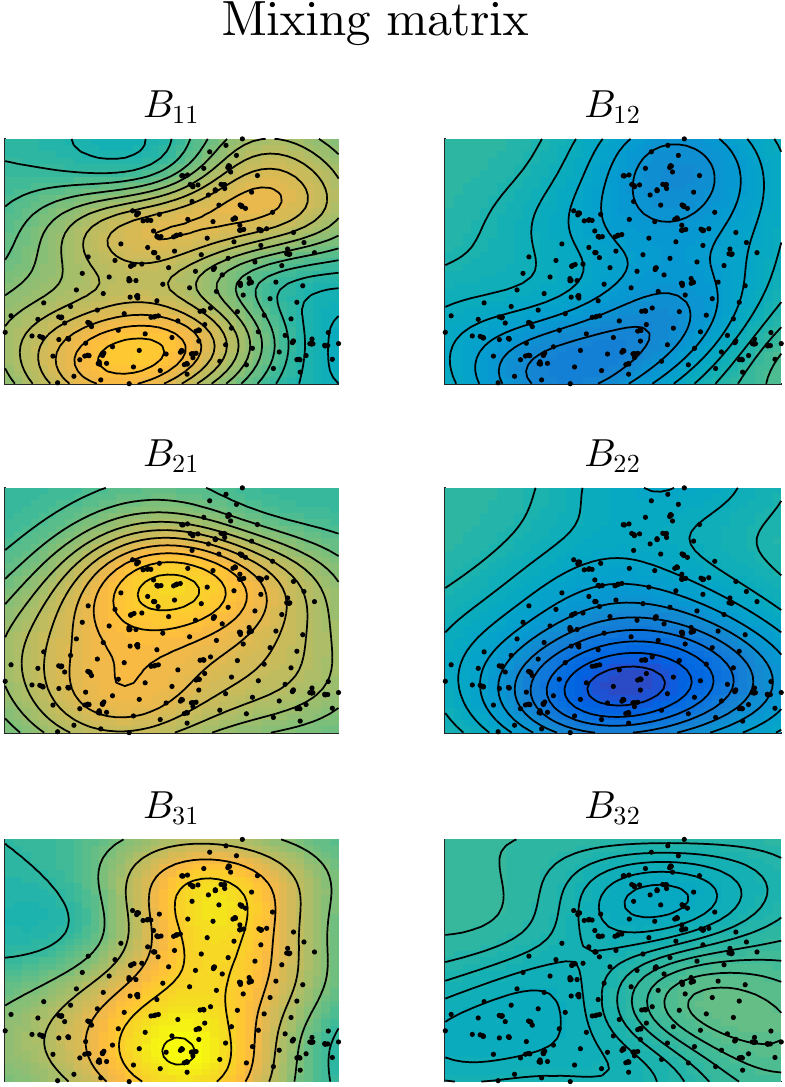}}
    \hfill
    \subfigure[$\y(x) \approx \f(\x) = \B(\x)\u(\x)$]{\includegraphics[height=2.2in]{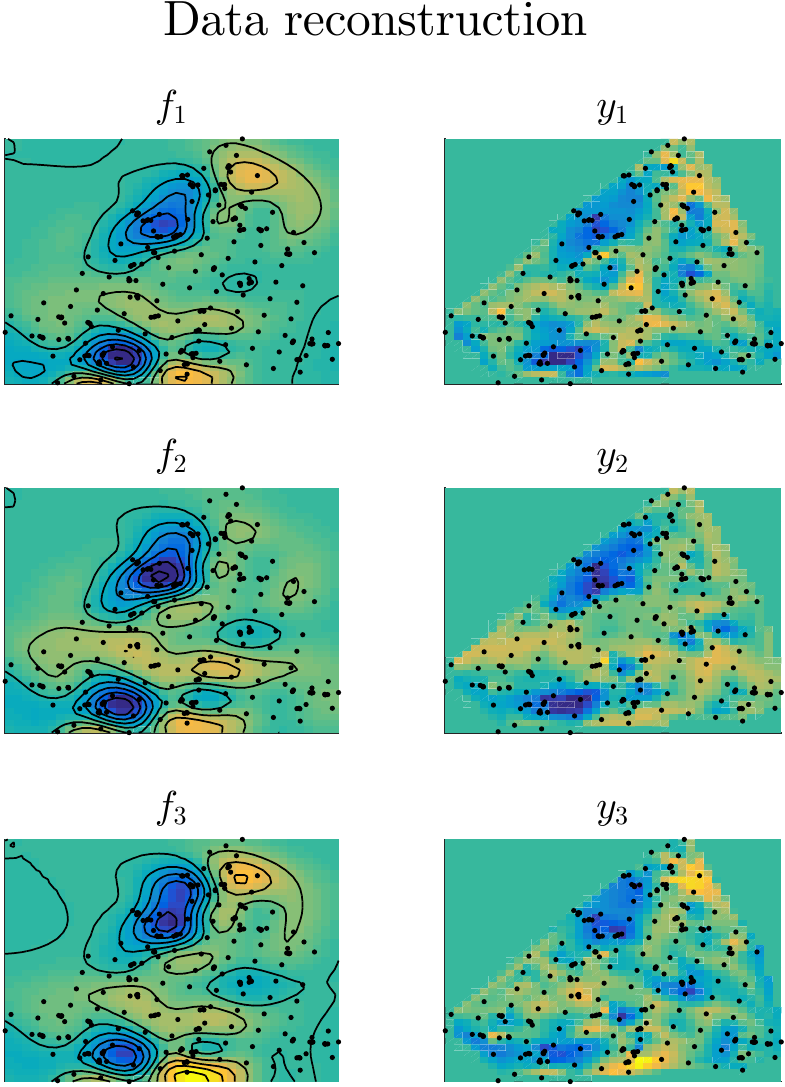}}
    \caption{Results on the Jura data. \textbf{(a)}: The learned latent covariance matrix shows an overall negative correlation over the input space. \textbf{(b)}: The learned latent variables clearly show a negative correlation. \textbf{(c)}: The learned input-dependent mixing matrix. \textbf{(d)}: LCGP reconstructs accurately the original data, which is plotted using a simple linear interpolation of the observed points for comparison.}
    \label{fig:jura}
\end{figure}

The Jura dataset\footnote{Data available at \url{https://sites.google.com/site/goovaertspierre/pierregoovaertswebsite/publications/book}.} 
consists of measurements of cadmium, nickel and zinc concentrations in a region of the Swiss Jura \citep{goovaerts1997}.
For training we are given the concentrations measured at 259 locations and for validation the measurements at 100 additional locations.
We set hyperparameters for both our model and GPRN as $\ell_u = 0.5$ and $\ell_b = 1$, 
and for our model the parameter for the latent correlation lengthscale to $\ell_z = 1$.
We learned the models with $Q=2$ latent variables, which resulted in the best model performance. We report both the mean squared and absolute errors for the predicted concentrations 
in Table~\ref{tab:results}. Our model performs at the same level as the state-of-the-art competitor GPRN, with slightly better performance in absolute errors.

Figure~\ref{fig:jura} shows the inferred model. The latent variables are 2D spatial surfaces on which the measurement points are indicated as black points. The two latent variables learn different geological processes that have an interesting two-pronged correlation pattern that indicates two kinds of negative correlations (the scatter plot). By explicitly modelling the latent covariance, we are able to see the regions of the input space that contribute to this pattern; the latent covariances indicate the combined covariance $C_{pq}(\x,\x') = A_{\x \x'}(p,q) K_{pq}(\x, \x')$. The diagonal plots of Figure~\ref{fig:jura}a show the variances of the two latent signals, while the off-diagonal covariance plot indicates the two-pronged negative correlation model between the geological processes. Finally, the mixing matrices of the two latent components reconstruct the three ore observation surfaces.

\subsection{EEG}

\begin{figure}[t]
    \centering
    \begin{minipage}{\textwidth}
    \subfigure[$\Z\Z^T \circ \K_Q$]{\includegraphics[height=1.9in]{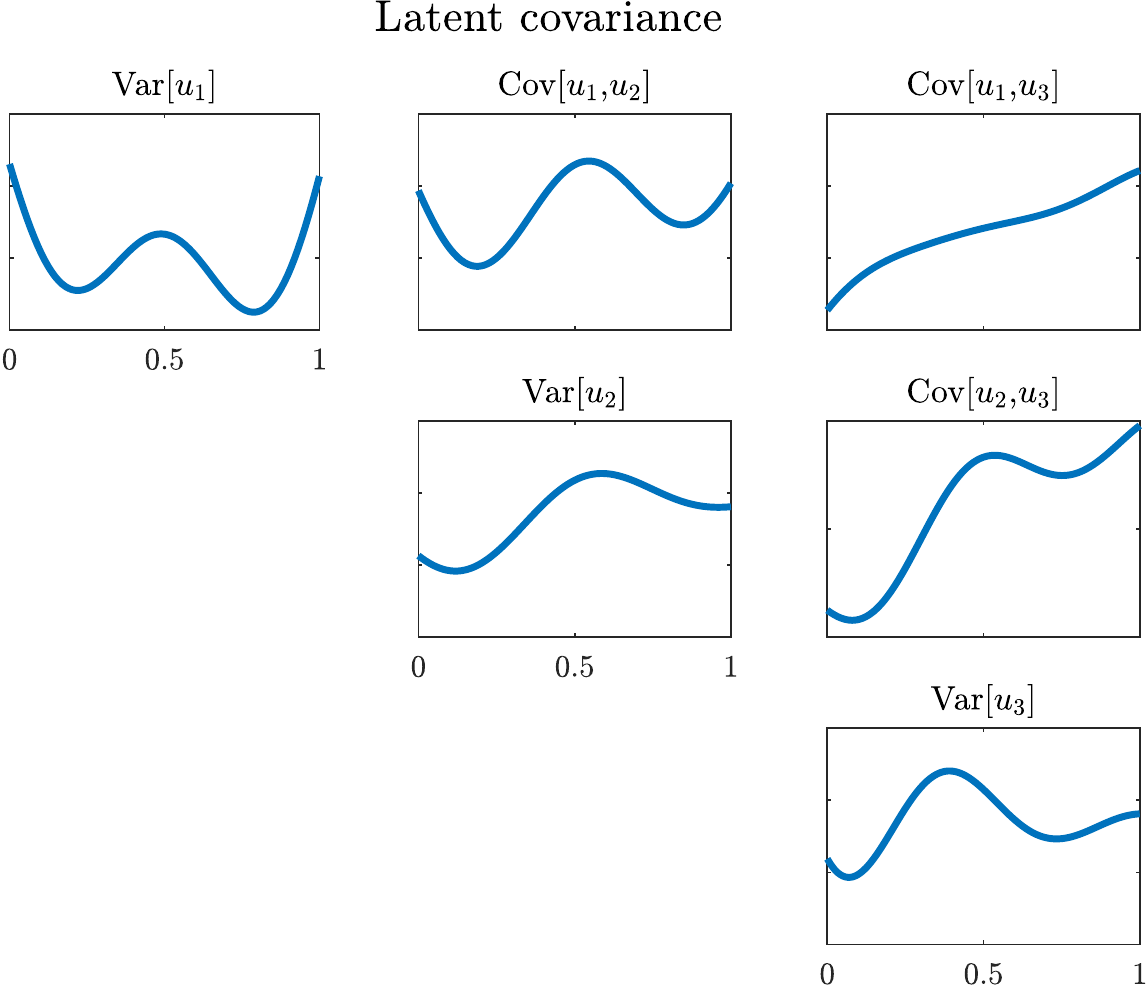}}
    \subfigure[$\u \sim \N(\vec0,\Z\Z^T\circ\K_Q)$]{\includegraphics[height=1.9in]{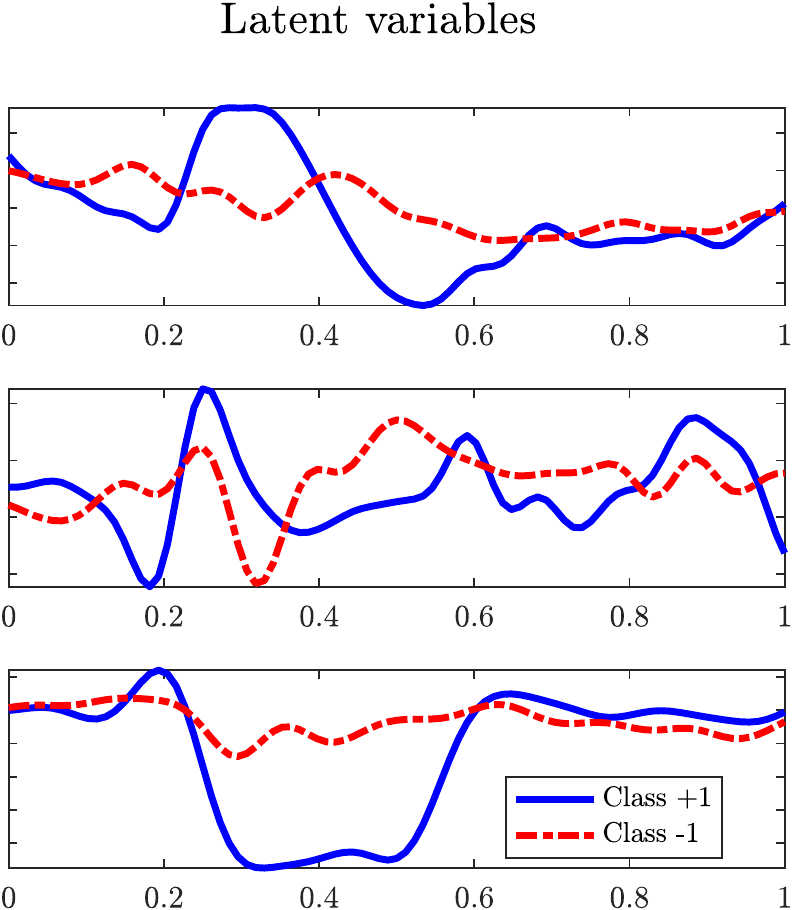}}
    \subfigure[$\y \sim \N(\B\u,\omega_f^{-1}\vec I)$]{\includegraphics[height=1.9in]{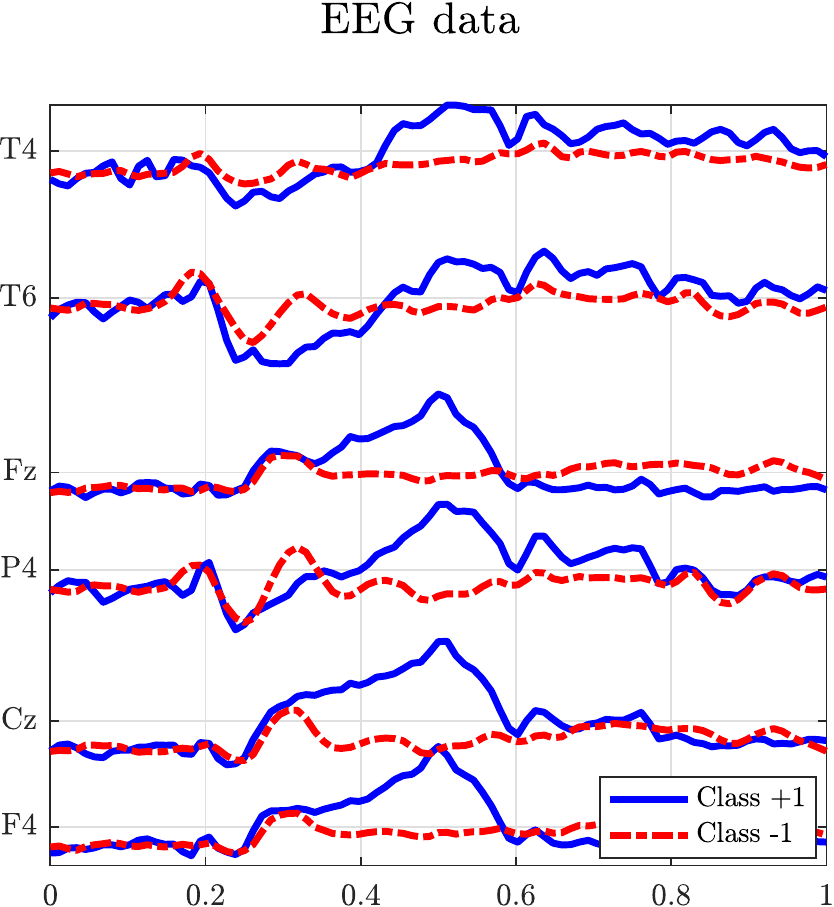}}
    \end{minipage}
    \centering
    \begin{minipage}{\textwidth}
    \subfigure[]{\begin{minipage}{\linewidth}\includegraphics[width=.32\linewidth]{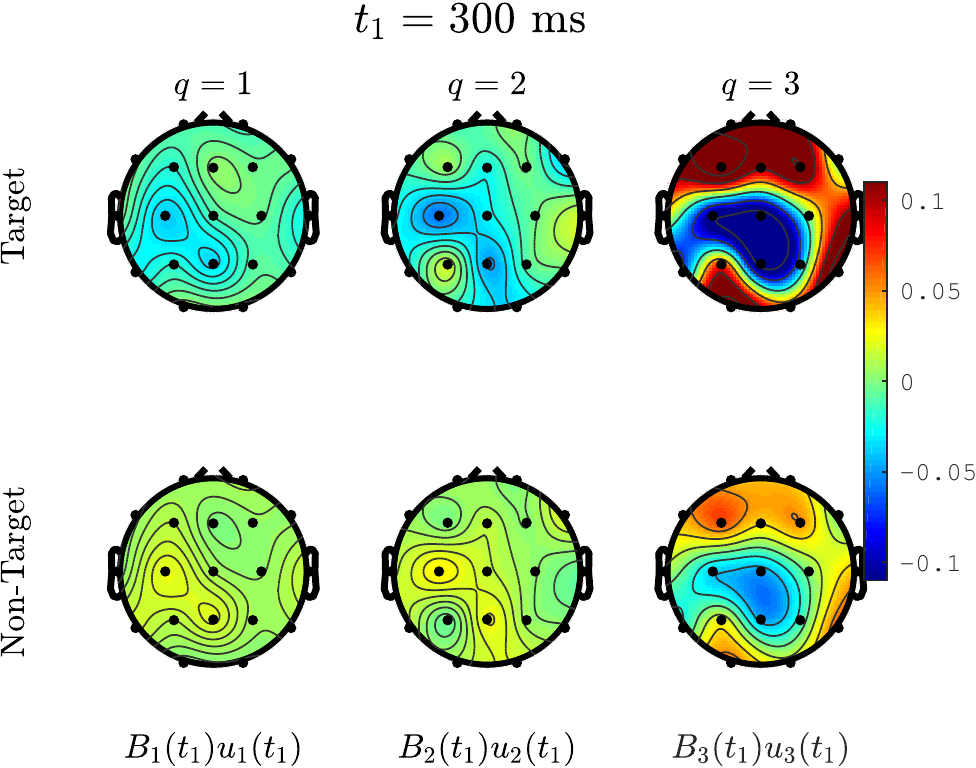}
    \includegraphics[width=.32\linewidth]{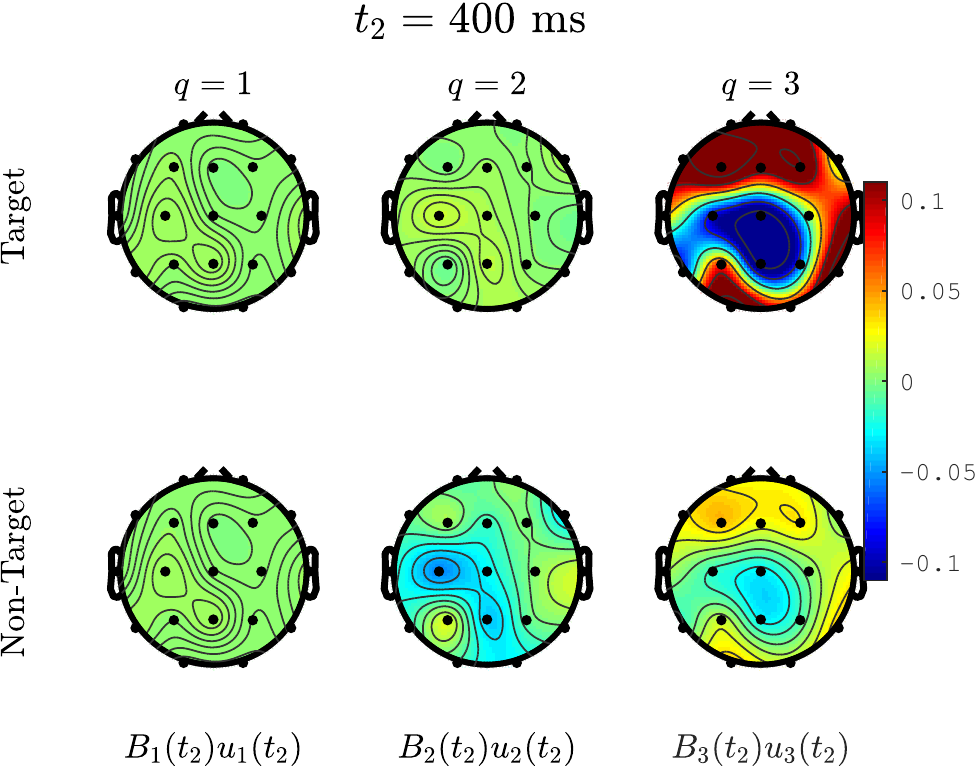}
    \includegraphics[width=.32\linewidth]{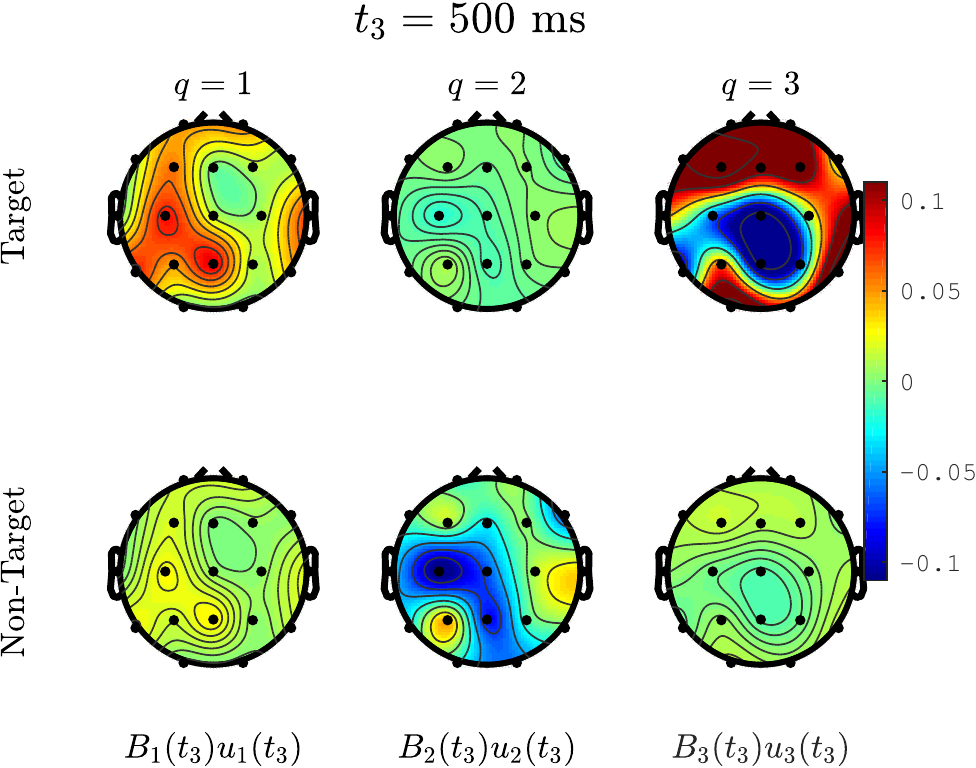}\end{minipage}}
    \end{minipage}
    \caption{\textbf{(a)}: Latent covariances. \textbf{(b)}: Latent variables, averaged within each class. \textbf{(c)} Class-wise averaged EEG data from a subset of 6 channels. \textbf{(d)} LCGP decomposes the brain activity to several components, showing the progression over the EEG scalp map at several time points $t$. The top figures shows the target class (+1) and bottom the non-target (-1), for each $t$.}
    \label{fig:eeg}
\end{figure}

Our main motivation for developing the present model was in modelling EEG data. 
We demonstrate LCGP on data from a P300 study \citep{vareka2014event}, where
the subjects were shown either a target or non-target stimulus, specifically a green or a red LED 
flashing, respectively. The classification task is to classify the stimulus based on the brain measurements.
Additionally, we evaluate our modelling approach in a simulation study.

\subsubsection{Classification Results}

We evaluate the classification performance using a Monte Carlo cross-validation scheme where in each fold we randomly sample training and test sets of $S=1000$ trials from the full dataset consisting in total of 7351 trials from 16 subjects. A single trial is the continuous voltage measurement of $M=19$ channels in an EEG cap for 800ms with $N=89$ after filtering and downsampling the time series \citep{hoffmann2008efficient}. We report the average area under the ROC curve (AUC) statistic over 100 folds, and compare our method to the state-of-the-art regularised LDA method implemented in the BBCI toolbox \citep{blankertz2010berlin}. Results in Table~\ref{tab:results} show that our method performs better than RLDA ($p<0.05$).

An example visualization of the model from one of the cross-validation folds is depicted in Figure~\ref{fig:eeg} for the three first latent signals. Panel (a) indicates the shared variances and covariances of the latent signals along time. The first and third latent signals have a monotonically increasing covariance coupling, while the first and second latent signals have a periodicity in the covariance. The average latent variables of the target and non-target trials are shown in panel (b). The third latent variable captures a strong dynamic between time points $[0.3, 0.5]$, which coincides with the expected P300 activity approximately 300 ms after the stimulus representation. The first two variables show peaks also at approximately 300 ms. 
In general the positive trials have a remarkably different latent representations than the negative trials. Panel (c) shows the classifier weights $\w_p$ for the three latent signals with the average classification plotted. Finally, a subset of the EEG channels are shown in panel (d), highlighting the differences in the channel dynamics. In panel (e) the components are found to be discriminative also when plotted on the scalp map.

\subsubsection{Finding the Latent Correlations}

In addition to the classification results, we evaluated our modelling approach in a simulation study to test whether we can find the latent correlations correctly using data that resembles the real EEG as closely as possible, but where we know the ground truth. To this end, we simulated datasets from the fitted models, and learn a new model on the simulated data.
We repeated this for varying number of latent variables ($Q = 2,3,4$) with 10 simulations done for each value of $Q$. For each simulation we computed the empirical $p$-value with the hypothesis that the accuracy of our model is greater than using randomly simulated covariances from the model. The $p$-values from the simulations were combined using the Fisher's method \citep{fisher1925}, with results reported in Table~\ref{tab:results}. We use again the correlation-based score for assessing the accuracy as with the toy data case.

\section{Discussion}

The LCGP is a flexible framework for multi-task learning. We demonstrate in two experiments that our model can robustly learn the latent variable correlations. The model also achieves state-of-the-art performance in both regression and classification. The added modeling of the correlations of the underlying latent processes both improves model interpretability, and regularises the model especially with multiple observations. The novel Wishart-Gibbs cross-covariance kernel encodes mutually-dependent covariances between latent signals and inputs in a parameterised way without being too flexible.

In place of the non-stationary Gaussian kernel other non-stationary kernels are possible. \citet{paciorek2006} propose a class of non-stationary convolution kernels containing, for instance, a non-stationary Mat\'ern kernel. For future work coupling the spectral kernels \citep{wilson2013} with Wishart correlations is another highly interesting avenue for a general family of dependent, structured kernels. The mutually-dependent Hadamard kernel would also be interesting to study in context of structured multi-task learning to model dependent input-output relations.

\acks{{\small This work was supported by the Finnish Funding Agency for Innovation (project Re:Know) and Academy of Finland (COIN CoE, and grants 299915, 294238 and 292334). The research leading to this results has received funding from the European Union Seventh Framework Programme (FP7/2007-2013) under grant agreement no 611570. We acknowledge the computational resources provided by the Aalto Science-IT project.\par}}
\bibliography{refs}

\appendix

\section{Optimisation of kernel parameters}

The factor $q(\Z)$ corresponding to the GP parameters of the proposed Hadamard kernel is updated by finding point estimates by maximizing the variational lower bound. The relevant part of the bound is given by
\begin{align*}
    \L(\Z) &=  \sum_s \langle \log p(\u^{(s)}|\Z) \rangle + \log p(\Z)
    = -\half(S\log|\K| + \sum_s \langle \u^{(s)T}\K^{-1}\u^{(s)} \rangle + \Tr ( \Z^T\K_z^{-1}\Z )) \label{eq:z-cost}
\end{align*}
where $\K = \Z\Z^T \circ \K_Q + \bO$, and its gradient by
\begin{align*}
    \pdiff{\L}{\Z_{ij}} = \half \Tr\left( \left[ \K^{-1}(\sum_s \langle \u^{(s)}\u^{(s)T} \rangle)\K^{-1} - S\K^{-1} \right] \pdiff{\K}{\Z_{ij}} \right) - \half [\K_z^{-1}\Z]_{ij} 
\end{align*}
where  $\K_z = K_z \otimes I_Q$ is a block matrix of full size $(QN \times QN$), and
\begin{align*}
    \pdiff{\K}{\Z_{ij}} = \pdiff{(\Z\Z^T \circ \K_Q + \bO)}{\Z_{ij}} &= (\Z\vec1_{ij}^T + \vec1_{ij}\Z^T) \circ \K_Q.
\end{align*}
The cost function in the whitened domain can be evaluated as $\L(\vec L \hat\Z)$ and gradient as 
\[ \pdiff{\L}{\hat{\Z}} = \pdiff{L}{\Z} \pdiff{\Z}{\hat{\Z}} = \vec L^T \pdiff{\L}{\Z}\,. \]
We can similarly optimize noise precision $\omega_u$ and lengthscales $\ell_u$.

\end{document}